%% file: aaai22.tex
\def\year{2022}\relax
\documentclass[letterpaper]{article} 
\usepackage{aaai22}  
\usepackage{times}  
\usepackage{helvet}  
\usepackage{courier}  
\usepackage[hyphens]{url}  
\usepackage{graphicx} 
\usepackage{natbib}  
\usepackage{caption} 
\DeclareCaptionStyle{ruled}{labelfont=normalfont,labelsep=colon,strut=off} 
\usepackage{algorithm}
\usepackage{algorithmic}
\usepackage{booktabs}
\usepackage[switch]{lineno}

\usepackage{newfloat}
\usepackage{listings}
\usepackage{amssymb}
\usepackage{amsmath}
\usepackage{multirow}
\usepackage{subcaption}
\floatstyle{ruled}
\newfloat{listing}{tb}{lst}{}
\floatname{listing}{Listing}
\title{SOIT: Segmenting Objects with Instance-Aware Transformers}

\author {
    Xiaodong Yu\textsuperscript{\rm 1}\thanks{Equal contribution. \textsuperscript{$\dagger$}Corresponding author.},
    Dahu Shi\textsuperscript{\rm 1}\footnotemark[1],
    Xing Wei\textsuperscript{\rm 2},
    Ye Ren\textsuperscript{\rm 1},
    Tingqun Ye\textsuperscript{\rm 1},
    Wenming Tan\textsuperscript{\rm 1\scriptsize{$\dagger$}}
}

\affiliations{
    \textsuperscript{\rm 1} Hikvision Research Institute, Hangzhou, China \\
    \textsuperscript{\rm 2} School of Software Engineering, Xi'an Jiaotong University \\
    \{yuxiaodong7, shidahu\}@hikvision.com,
    weixing@mail.xjtu.edu.cn, \\
    \{renye, yetingqun, tanwenming\}@hikvision.com
}

\usepackage{bibentry}

\begin{document}

\maketitle

\input{parts/0_abstract}
\input{parts/1_intro}
\input{parts/2_related_work}
\input{parts/3_method}
\input{parts/4_experiment}
\input{parts/5_conclusion}

\bibliography{aaai22.bib}
\end{document}

%% file: parts/0_abstract.tex
\begin{abstract}

This paper presents an end-to-end instance segmentation framework, termed SOIT, that Segments Objects with Instance-aware Transformers.
Inspired by DETR~\cite{carion2020end}, our method views instance segmentation as a direct set prediction problem and effectively removes the need for many hand-crafted components like RoI cropping, one-to-many label assignment, and non-maximum suppression (NMS).
In SOIT, multiple queries are learned to directly reason a set of object embeddings of semantic category, bounding-box location, and pixel-wise mask in parallel under the global image context.
The class and bounding-box can be easily embedded by a fixed-length vector.
The pixel-wise mask, especially, is embedded by a group of parameters to construct a lightweight instance-aware transformer.
Afterward, a full-resolution mask is produced by the instance-aware transformer without involving any RoI-based operation.
Overall, SOIT introduces a simple single-stage instance segmentation framework that is both RoI- and NMS-free.
Experimental results on the MS COCO dataset demonstrate that SOIT outperforms state-of-the-art instance segmentation approaches significantly.
Moreover, the joint learning of multiple tasks in a unified query embedding can also substantially improve the detection performance.
Code is available at \url{https://github.com/yuxiaodongHRI/SOIT}.

\end{abstract}

%% file: parts/1_intro.tex
\begin{figure}[t]
    \centering
    \subfloat[Detect-then-segment pipeline.]
    {
        \includegraphics[width=0.9\columnwidth]{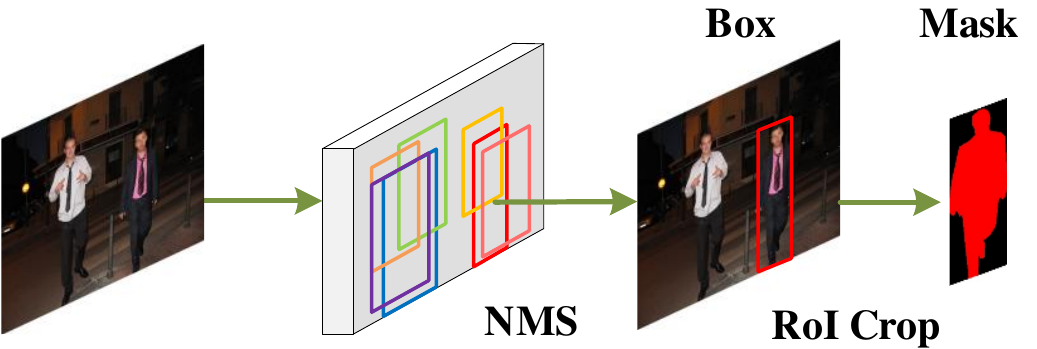}
        \label{fig:pipelines:two-stage}
    }
    \newline
    \subfloat[Detect-and-segment pipeline.]
    {
        \includegraphics[width=0.9\columnwidth]{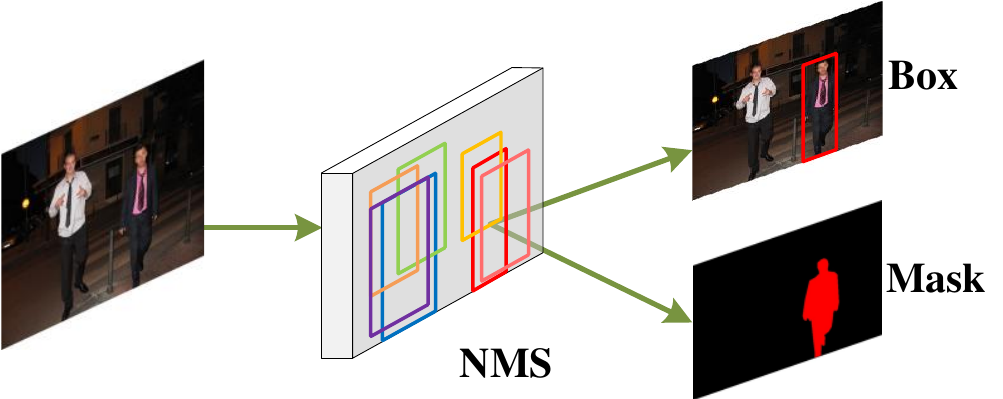}
        \label{fig:pipelines:single-stage}
    }
    \newline
    \subfloat[Fully end-to-end pipeline.]
    {
        \includegraphics[width=0.9\columnwidth]{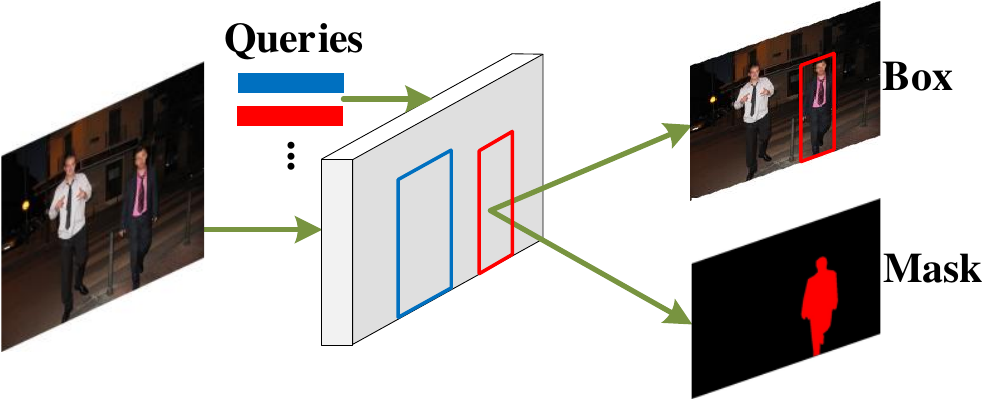}
        \label{fig:pipelines:e2e}
    }
    \caption{Comparisons of different instance-level perception pipelines. We proposed the fully end-to-end framework as shown in (c), which is RoI-free and NMS-free.} 
    \label{fig:pipelines}
\end{figure}

\section{Introduction}
\label{sec:intro}

Instance segmentation is a fundamental yet challenging task in computer vision, which requires an algorithm to predict a pixel-wise mask with a category label for each instance of interest in an image.
As popularized in the Mask R-CNN framework~\cite{he2017mask}, state-of-the-art instance segmentation methods follow a \emph{detect-then-segment} paradigm \cite{cai2019cascade, chen2019hybrid, vu2021scnet}.
These methods employ an object detector to produce the bounding boxes of instances and crop the feature maps via RoIAlign \cite{he2017mask} according to the detected boxes.
Then pixel-wise masks are predicted by a fully convolutional network (FCN) \cite{long2015fully} only in the detected region (as shown in Fig \ref{fig:pipelines:two-stage}).
The detect-then-segment paradigm is sub-optimal since it has the following drawbacks: 1) Segmentation results heavily rely on the object detector, incurring inferior performance in complex scenarios; 2) RoIs are always resized into patches of the same size (\emph{e.g.}, $14 \times 14$ in Mask R-CNN), which restricts the quality of segmentation masks, as large instances would require higher resolution features to retain details at the boundary.
To overcome the drawbacks of this paradigm, recent works \cite{chen2019tensormask, xie2020polarmask, cao2020sipmask, peng2020deep} start to build instance segmentation frameworks on top of single-stage detectors \cite{lin2017focal, tian2019fcos}, getting rid of local RoI operations.
However, these methods still rely on one-to-many label assignment in training and hand-crafted Non-Maximum Suppression (NMS) post-processing to eliminate duplicated instances when testing.
As a result, these two categories of instance segmentation methods are not end-to-end fully optimized and suffer from sub-optimal solutions.

Inspired by the recent application of transformer architecture in object detection \cite{carion2020end, zhu2021deformable}, we present a transformer-based instance segmentation framework, namely SOIT (Segment Objects with Instance-aware Transformer) in this paper.
We reformulate instance segmentation as a direct set prediction problem and builds a fully end-to-end approach.
Concretely, given multiple randomly initialized object queries, SOIT learns to reason a set of object embeddings of semantic category, bounding-box, and pixel-wise mask simultaneously, under the global image context.
SOIT adopts the bipartite matching strategy to assign a learning target for each object query.
As shown in Fig. \ref{fig:pipelines:e2e}, this training approach is advantageous to conventional one-to-many instance segmentation training strategies \cite{he2017mask, wang2020solo, tian2020conditional} as it avoids the heuristic label assignment and eliminates the need for NMS post-processing.

A compact fixed-length vector can easily embed the semantic category and bounding-box in the end-to-end learning framework.
However, it is not trivial to represent a spatial binary mask of each object for learning as the mask is high-dimensional and varies from each instance. 
To solve this problem, we embed the pixel-wise mask to a group of instance-aware parameters, whereby a unique instance-aware transformer is constructed.
Moreover, we propose a novel relative positional encoding for the transformer, which provides strong location cues to distinguish different objects.
Then the instance-aware transformer is employed to segment the object in a high-resolution feature map directly.
It is expected that the instance-aware parameters and relative positional encoding can encode the characteristics of each instance. Thus it can only fire on the pixels of the particular object.
As described above, our method is naturally RoI-free and NMS-free, which eliminates many extra hand-crafted operations involved in previous instance segmentation methods.

Our main contributions are summarized as follows:

\begin{itemize}

\item
We attempt to solve instance segmentation from a new perspective that uses parallel instance-aware transformers in an end-to-end framework.
This novel solution enables the framework to directly generate pixel-wise mask results of each instance without RoI cropping or NMS post-processing.

\item 
In our method, queries learn to encode multiple object representations simultaneously, including categories, locations, and pixel-wise masks.
This multi-task joint learning paradigm establishes a collaboration between objection detection and instance segmentation, encouraging these two tasks to benefit from each other.
We demonstrate that our architecture can also significantly improve object detection performance.

\item
To show the effectiveness of the proposed framework, we conduct extensive experiments on the COCO dataset.
SOIT with ResNet-50 achieves 42.5\% mask AP and 49.1\% box AP on the \texttt{test-dev} split without any bells and whistles, outperforming the complex well-tuned HTC \cite{chen2019hybrid} by 2.8\% in mask AP and 4.2\% in box AP.

\end{itemize}

%% file: parts/2_related_work.tex
\section{Related Work}
\label{sec:relwork}

\subsection{Instance Segmentation}
Instance segmentation is a challenging task, as it requires instance-level and pixel-wise predictions simultaneously.
The existing approaches can be summarized into three categories: top-down, bottom-up, and single-stage methods.
In top-down methods, the Mask R-CNN family \cite{he2017mask, cai2019cascade, chen2019hybrid, cao2020d2det} follow the detect-then-segment paradigm, which first performs object detection and then segments objects in the boxes.
Moreover, some recent works \cite{lee2020centermask, wang2020rdsnet, chen2020supervised} are proposed to improve the segmentation performance further.
Bottom-up methods \cite{liu2017sgn, gao2019ssap} view the task as a label-then-cluster problem. They first learn per-pixel embeddings and then cluster them into instance groups.
Besides, YOLACT \cite{bolya2019yolact}, CondInst \cite{tian2020conditional} and SOLO \cite{wang2020solo} build single-stage instance segmentation framework on the top of one-stage detectors \cite{tian2019fcos}, achieving competitive performance.
Concurrently, QueryInst~\cite{Fang_2021_ICCV} and SOLQ~\cite{dong2021solq} aim at building end-to-end instance segmentation frameworks, eliminating NMS post-processing.
However, they still need RoI cropping to separate different instances first, which may have the same limitations of the detect-then-segment pipeline.
In this paper, we go for an end-to-end instance segmentation framework that neither relies on RoI cropping nor NMS post-processing.

\subsection{Transformer in Vision}
Transformer \cite{vaswani2017attention} introduces the self-attention mechanism to model long-range dependencies and has been widely applied in natural language processing (NLP). Recently, several works attempted to involve Transformer architecture in computer vision tasks and showed promising performances. 
ViT series \cite{dosovitskiy2020image, touvron2021training} take an image as a sequence of patches and achieve the cross-patch interactions by transformer architecture in image classification.
DETR \cite{carion2020end}, and Deformable DETR \cite{zhu2021deformable} adopted learnable queries and transformer architecture together with bipartite matching to perform object detection in an end-to-end fashion, without any hand-crafted process such as NMS.
SETR \cite{zheng2021rethinking} reformulates the image semantic segmentation problem from a sequence-to-sequence learning perspective, offering an alternative to the dominating encoder-decoder FCN model design.
Despite transformer architecture is being widely used in many computer vision tasks, few efforts are conducted to build a transformer-based instance segmentation framework. We aim to achieve this goal in this paper.

\begin{figure*}[t]
  \centering
  \includegraphics[width=.95\textwidth]{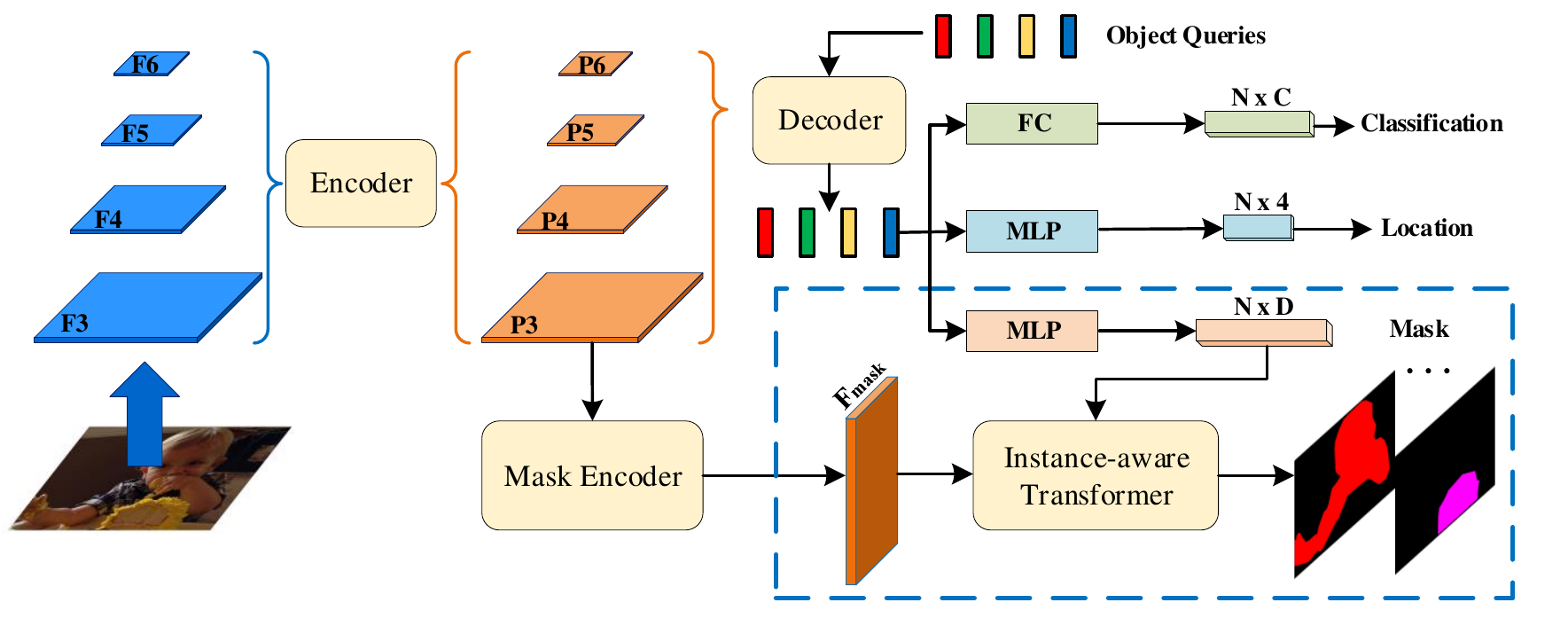}
  \caption{Illustration of the overall architecture of SOIT. $F_3$ to $F_6$ are the multi-scale image feature maps extracted from the backbone (\emph{e.g.}, ResNet-50). $P_3$ to $P_6$ are the multi-scale feature memory refined by the transformer encoder. $F_{mask}$ represents mask features produced by the mask encoder. D-dimensional (\emph{e.g.} 441 by default) dynamic parameters generated in the mask branch are used to construct the instance-aware transformer. As shown in the blue dashed box, the pixel-wise mask is produced via the instance-aware transformer, of which the details are described in Figure \ref{fig:instance-aware-transformer}.}
  \label{fig:overall-architecture}
\end{figure*}

\subsection{Dynamic Networks}
Unlike traditional network layers with fixed filters once trained, the filters of dynamic networks are conditioned on the input and dynamically generated by another network. 
This idea has been explored previously in convolution modules like dynamic filter networks \cite{jia2016dynamic}, and CondConv \cite{yang2019condconv}, to increase the capacity of a classification network.
Recently, some works \cite{tian2020conditional, shi2021inspose} employ the dynamic filters, conditioned on each instance in the image, to implement instance-level vision tasks.
In this work, we extend this idea to transformer architecture and build instance-aware transformers to solve the challenging instance segmentation task.

%% file: parts/3_method.tex
\section{Method}

In this section, we first introduce the overall architecture of our framework. Next, we elaborate on the proposed instance-aware transformer employed to produce the full-resolution mask for each instance. Then, we describe relative positional encoding to improve instance segmentation performance further. At last, the training losses of our model are summarized.

\subsection{Overall Architecture}
As depicted in Fig. \ref{fig:overall-architecture}, the proposed framework is composed of three main components: a backbone network to extract multi-scale image feature maps, a transformer encoder-decoder to produce object-related query features in parallel, and a multi-task prediction network to perform object detection and instance segmentation simultaneously.

\subsubsection{Multi-Level Features.}
Given an image $I \in \mathbb{R}^{H \times W \times 3}$, we extract multi-scale feature maps $\boldsymbol{\mathcal{F}}=\{F_3, F_4, F_5, F_6\}$ (blue feature maps in Fig \ref{fig:overall-architecture}) from the backbone (\emph{e.g.}, ResNet \cite{he2016deep}).
Specifically, $\{F^l\}_{l=3}^{5}$ are produced by adding a $1 \times 1$ convolution on the output feature maps of stage $C_3$ through $C_5$ in the backbone, where $C_l$ is of resolution $2^l$ lower than the input images.
The lowest resolution feature map $F_6$ is obtained via a $3 \times 3$ stride 2 convolution on the final $C_5$ stage.
Multi-scale image feature maps in $\boldsymbol{\mathcal{F}}$ are all of 256 channels.

\subsubsection{Transformer Encoder-Decoder.}
In this work, we employ the deformable transformer encoder \cite{zhu2021deformable} to produce multi-scale feature memory.
Each encoder layer comprises a multi-scale deformable attention module \cite{zhu2021deformable} and a feed-forward network (FFN).
There are six encoder layers stacked in sequence in our framework.
The encoder takes the image feature maps $\boldsymbol{\mathcal{F}}$ as input and output the refined multi-scale feature memory $\boldsymbol{\mathcal{P}}=\{P^l\}_{l=3}^{6}$ (orange feature maps in Fig. \ref{fig:overall-architecture}) with the same resolutions.

Given the refined multi-scale feature memory $\boldsymbol{\mathcal{P}}$ and $N$ learnable object queries, we then generate the instance-aware query embeddings for target objects by the deformable transformer decoder \cite{zhu2021deformable}.
Similar to the encoder, six decoder layers are applied sequentially.
Each one is composed of a self-attention module, and a deformable cross-attention module \cite{zhu2021deformable}, where object queries interact with each other and the global context, respectively.
In the end, the instance-aware query features are collected and then fed into the multi-task prediction network.

\subsubsection{Multi-Task Predictions.}
After query feature extraction, each query embedding represents the features of the corresponding instance.
Subsequently, we simultaneously apply three branches to generate the category, bounding-box location, and pixel-wise mask of the targeting instance.
The classification branch is a linear projection layer (FC) that predicts the class confidence for each object.
The location branch is a multi-layer perceptron (MLP) with a hidden size of 256 and predicts the normalized center coordinates, height, and width of the box \emph{w.r.t.} the input image.
The mask branch architecture is the same as the location branch except that the channel of the output layer is set to $D$.
It is worth noting that the output of the mask branch is a group of dynamic parameters conditioned on the particular instance.
These parameters are later employed to construct instance-aware transformers to directly generate masks from full-image feature maps, elaborated in the following subsection.

\subsection{Instance-Aware Transformers}
Unlike semantic category and bounding-box, it is challenging to represent the per-pixel mask by a compact fixed-length vector without RoI cropping.
Our core idea is that for an image with $N$ instances, $N$ different transformer encoder networks will be dynamically generated.
It is expected that the instance-aware transformer can encode the characteristics of each instance and only fires on the pixels of the corresponding object.
To avoid the quadratic growth of the computational complexity in the original transformer encoder \cite{vaswani2017attention}, we build our instance-aware transformer on the deformable transformer encoder \cite{zhu2021deformable} for efficiency.

\begin{figure}[t]
    \centering
    \includegraphics[width=1.0\columnwidth]{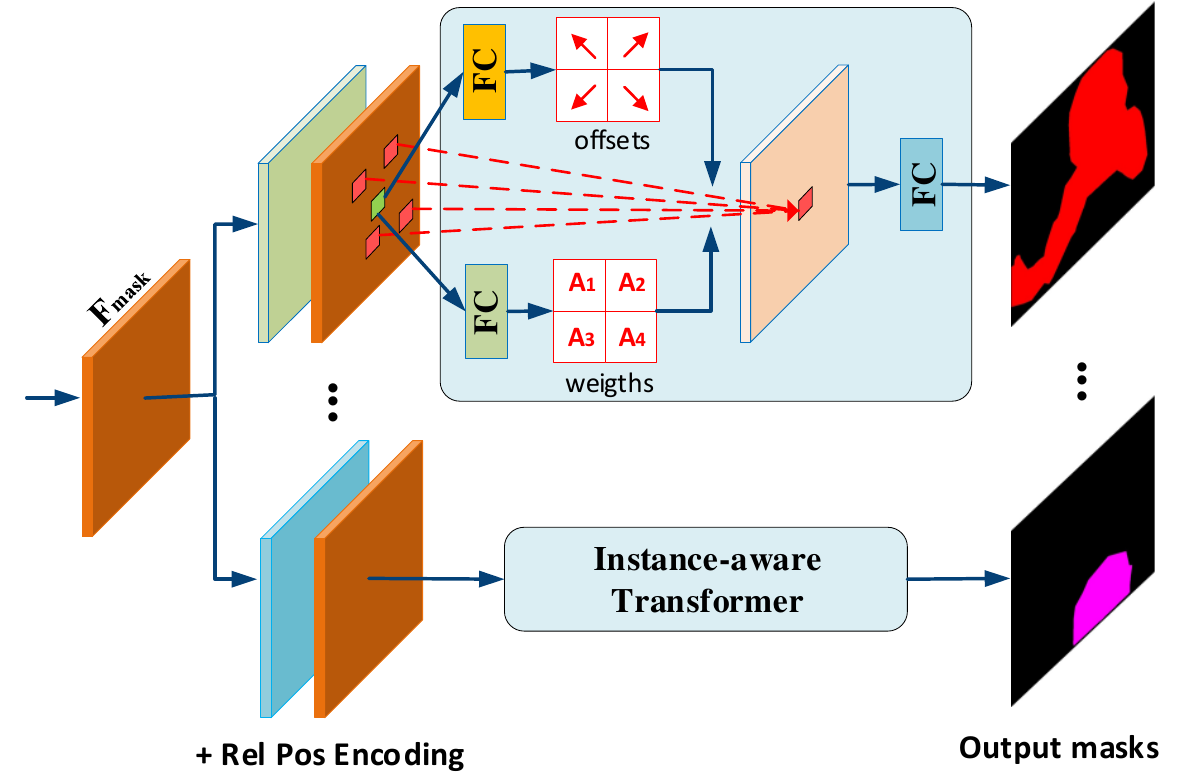}
    \caption{Detailed structure of the instance-aware transformer. Two linear projections (\emph{i.e.}, FC) predict sampling locations and attention weights for different feature points, respectively. Another linear projection is employed for output projection. In our instance-aware transformer, all weights of these three layers are dynamically generated in the mask branch and conditioned on the target object.}
    \label{fig:instance-aware-transformer}
\end{figure}

Concretely, given an input feature map $\boldsymbol{x} \in \mathbb{R}^{C \times H \times W}$, let $q$ indexes a query (\emph{e.g.}, the green grid point in Fig. \ref{fig:instance-aware-transformer}) with content feature $\boldsymbol{z}_q$ and a 2-d reference point $p_q$, the deformable multi-head attention feature is calculated by
\begin{equation}
    H_m^n = \sum_{k=1}^{K}A_{mqk}^n \cdot \boldsymbol{x}(p_q + \Delta p_{mqk}^n),
\end{equation}
where $m \in [1, 2, \ldots, M]$ indexes the attention head, $k$ indexes the sampled keys, and $K$ is the total sampled key number ($K \ll HW $). $n$ denotes the $n$-th object query (\emph{i.e.}, instance).
As shown in Fig. \ref{fig:instance-aware-transformer}, $\Delta p_{mqk}$ and $A_{mqk}$ are the sampling offset and attention weight of the $k^{\text{th}}$ sampling point in the $m^{\text{th}}$ attention head, respectively. Both $\Delta p_{mqk}$ and $A_{mqk}$ are obtained via a linear projection (\emph{i.e.}, FC) layer over the query feature $\boldsymbol{z}_q$. Afterwards, another linear projection layer (\emph{i.e.}, $W^n$) is applied for output projection, which can be formulated as
\begin{equation}
    \text{Mask}^n = W^n \left[ \text{Concat}\left(H_1^n, H_2^n, \ldots, H_M^n \right) \right],
\end{equation}
where ``Concat'' represents the concatenating operation.
To establish our instance-aware transformer encoder, the weights of these three linear projection layers are dynamically generated, conditioned on the target instance.
Specifically, for $n$-th object query, the $D$ parameters predicted in the mask branch are split into three parts and converted as the weights of the three linear projections.
Moreover, the channel of the output projection layer is set to 1 for the mask prediction, followed by a sigmoid activate function.
Note that the attention locations and weights for each instance are different even at the same feature point, so each instance has a particular preference for where to focus in the feature map.

\textbf{Shared Mask Features.}\quad
To get high-quality masks, our method generates pixel-wise masks on a full-image feature map, not a cropped region with fixed size (\emph{e.g.}, $14 \times 14$ in Mask R-CNN \cite{he2017mask}).
As shown in Fig. \ref{fig:overall-architecture}, 
the mask encoder branch is employed to provide the high-resolution feature map $F_{mask} \in \mathbb{R}^{H_{mask} \times W_{mask} \times C_{mask}}$ that the instance-aware transformers take as input to predict the per-instance mask.
The mask encoder branch is connected to aggregated feature $P_3$, and thus, its output resolution is $1/8$ of the input image.
It consists of a deformable transformer encoder layer, whose feature dimension is 256 (same as the feature channels of $P_3$). 
Afterward, a linear projection layer with layer normalization (LN) is employed to reduce the feature dimension from 256 to 8 (\textit{i.e.}, $C_{mask}=8$).
As described above, the instance-aware transformer mask head is very compact due to the few channels of the shared mask feature.

\subsection{Relative Positional Encodings}
As described in \cite{vaswani2017attention}, the original positional encoding in transformer is calculated by sine and cosine functions of different frequencies:
\begin{equation}
    \begin{split}
    PE_{(pos,2i)}=sin(pos/10000^{2i/d_{model}}) \\
    PE_{(pos,2i+1)}=cos(pos/10000^{2i/d_{model}}) 
    \end{split}
    \label{AbsPosEncoding}
\end{equation}
where $pos$ is the absolute position, $i$ is the dimension and $d_{model}$ is the embedding dimension.
DETR \cite{carion2020end} extends the above positional encoding to the 2D case.
Specifically, for both spatial coordinates $(x, y)$ of each embedding in the 2D feature map, DETR independently uses $d_{model}/2$ sine and cosine functions with different frequencies. Then they are concatenated to get the final $d_{model}$ channel positional encoding.

For our instance-aware transformer encoder, the input is the sum of the shared mask feature and the absolute positional encoding as described above.
To further utilize the location information of each object query, we propose a new relative positional encoding, which can be written as:
\begin{equation}
\begin{split}
PE_{(pos,2i)}=sin((pos-pos_{q})/10000^{2i/d_{model}}) \\
PE_{(pos,2i+1)}=cos((pos-pos_{q})/10000^{2i/d_{model}})
\end{split} \label{RelPosEncoding}
\end{equation}
where $pos_{q}$ is the center location of the box predicted by current object query.
Please note that the proposed relative positional encoding provides a strong cue for predicting the instance mask.
The performance improvement in the ablation study demonstrates its superiority compared to the original absolute positional encoding.

\subsection{Training Loss}
In this work, the final outputs of our framework are supervised by three sub-tasks: classification, localization, and segmentation. 
We use the same loss functions for classification and localization as in \cite{zhu2021deformable}, and adopt the Dice Loss \cite{milletari2016v} and the binary cross-entropy (BCE) loss for instance segmentation. 
The overall loss function is written as:
$$L=\lambda_{cls}L_{cls} + \lambda_{L1}L_{L1} + \lambda_{iou}L_{iou} + \lambda_{dice}L_{dice} + \lambda_{bce}L_{bce}.$$
Following \cite{zhu2021deformable}, we set $\lambda_{cls} = 2$, $\lambda_{L1} = 5$ and $\lambda_{iou} = 2$.
We empirically find $\lambda_{dice} = 8$ and $\lambda_{bce} = 2$ work best for the proposed framework.

%% file: parts/4_experiment.tex
\begin{table*}[t]
    \centering
    \resizebox{1.0\textwidth}{!}{
    \begin{tabular}{l|c|c|c|c|cc|ccc|c}
        \toprule
        Method & Backbone & RoI-free & NMS-free & AP & AP$_{50}$ & AP$_{75}$ & AP$_{S}$ &
         AP$_{M}$ & AP$_{L}$ & AP$^{box}$ \\
        \hline\hline
        Mask R-CNN \cite{he2017mask}          & \multirow{12}{*}{ResNet-50} & & & 37.5 & 59.3 & 40.2 & 21.1 & 39.6 & 48.3 & 41.3 \\
        CMR \cite{cai2019cascade}             & & & & 38.8 & 60.4 & 42.0 & 19.4 & 40.9 & 53.9 & 44.5 \\
        HTC \cite{chen2019hybrid}             & & & & 39.7 & 61.4 & 43.1 & 22.6 & 42.2 & 50.6 & 44.9 \\
        BlendMask \cite{chen2020blendmask}    & & & & 37.0 & 58.9 & 39.7 & 17.3 & 39.4 & 52.5 & 42.7 \\ 
        CondInst \cite{tian2020conditional}   & &\checkmark & & 37.8 & 59.2 & 40.4 & 18.2 & 40.3 & 52.7 & 41.9 \\ 
        SOLOv2 \cite{NEURIPS2020_cd3afef9}    & &\checkmark & & 38.2 & 59.3 & 40.9 & 16.0 & 41.2 & 55.4 & 40.4 \\ 
        DSC \cite{ding2021deeply}             & & & & 40.5 & 61.8 & 44.1 & -    & -    & -    & 46.0 \\
        RefineMask \cite{zhang2021refinemask} & & & & 40.2 & -    & -    & -    & -    & -    & -    \\
        SCNet \cite{vu2021scnet}              & & & & 40.2 & 62.3 & 43.4 & 22.4 & 42.8 & 53.4 & 45.0 \\
        SOLQ \cite{dong2021solq}              & & &\checkmark & 39.7 & - & - & 21.5 & 42.5 & 53.1 & 47.8 \\
        QueryInst \cite{Fang_2021_ICCV}       & & &\checkmark & 40.6 & 63.0 & 44.0 & 23.4 & 42.5 & 52.8 & 45.6 \\
        \textbf{SOIT} (Ours)                  & &\checkmark &\checkmark & \textbf{42.5} & \textbf{65.3} 
         & \textbf{46.0}  & \textbf{23.8} & \textbf{45.4} & \textbf{55.7} & \textbf{49.1} \\
        \hline \hline
        Mask R-CNN \cite{he2017mask}          & \multirow{14}{*}{ResNet-101} & & & 38.8 & 60.9 & 41.9 & 21.8 & 41.4 & 50.5 & 43.1 \\
        CMR \cite{cai2019cascade}             & & & & 39.9 & 61.6 & 43.3 & 19.8 & 42.1 & 55.7 & 45.7 \\
        HTC \cite{chen2019hybrid}             & & & & 40.7 & 62.7 & 44.2 & 23.1 & 43.4 & 52.7 & 46.2 \\
        MEInst \cite{zhang2020mask}           & & & & 33.9 & 56.2 & 35.4 & 19.8 & 36.1 & 42.3 & -    \\
        BlendMask \cite{chen2020blendmask}    & & & & 39.6 & 61.6 & 42.6 & 22.4 & 42.2 & 51.4 & 44.8 \\ 
        CondInst \cite{tian2020conditional}   & &\checkmark & & 39.1 & 60.9 & 42.0 & 21.5 & 41.7 & 50.9 & 43.3 \\ 
        SOLOv2 \cite{NEURIPS2020_cd3afef9}    & &\checkmark & & 39.7 & 60.7 & 42.9 & 17.3 & 42.9 & 57.4 & 42.6 \\
        DCT-Mask \cite{shen2021dct}           & & & & 40.1 & 61.2 & 43.6 & 22.7 & 42.7 & 51.8 & - \\
        DSC \cite{ding2021deeply}             & & & & 40.9 & 62.5 & 44.5 & -    & -    & -    & 46.7 \\
        RefineMask \cite{zhang2021refinemask} & & & & 41.2 & -    & -    & -    & -    & -    & -    \\
        SCNet \cite{vu2021scnet}              & & & & 41.3 & 63.9 & 44.8 & 22.7 & 44.1 & 55.2 & 46.4 \\
        SOLQ \cite{dong2021solq}              &  & &\checkmark & 40.9 & - & - & 22.5 & 43.8 & 54.6 & 48.7 \\
        QueryInst \cite{Fang_2021_ICCV}       &  & &\checkmark & 42.8 & 65.6 & 46.7 & 24.6 & 45.0 & 55.5 & 48.1 \\
        \textbf{SOIT} (Ours)                  & &\checkmark &\checkmark & \textbf{43.4} & \textbf{66.3} & \textbf{46.9} & \textbf{23.9} & \textbf{46.4} & \textbf{57.4} & \textbf{50.0} \\
        \hline \hline
        SOLQ \cite{dong2021solq}              & \multirow{3}{*}{Swin-L} & &\checkmark & 46.7 & - & - & 29.2 & 50.1 & 60.9 & 56.5 \\
        QueryInst \cite{Fang_2021_ICCV}       & & &\checkmark & 49.1 & 74.2 & \textbf{53.8} & \textbf{31.5} & 51.8 & 63.2 & 56.1 \\
        \textbf{SOIT} (Ours)                  & &\checkmark &\checkmark & \textbf{49.2} & \textbf{74.3} & 53.5 & 30.2 & \textbf{52.7} & \textbf{65.2} & \textbf{56.9} \\
        \bottomrule
    \end{tabular}
    }
    \caption{Comparisons with state-of-the-art instance segmentation methods on the COCO \texttt{test-dev}. ``CMR" is short for ``Cascade Mask RCNN". AP$^{box}$ denotes box AP, and AP without superscript denotes mask AP. All models are trained with multi-scale and tested with single scale.}
    \label{tab:comparison with SOTA}
\end{table*}

\section{Experiments}
\subsection{Dataset and Metrics}
We validate our method on COCO benchmark \cite{lin2014microsoft}. COCO 2017 dataset contains 115k images for training (split \texttt{train2017}), 5k for validation (split \texttt{val2017}) and 20k for testing (split \texttt{test-dev}), involving 80 object categories with instance-level segmentation annotations. Following the common practice, our models are trained with split \texttt{train2017}, and all the ablation experiments are evaluated on split \texttt{val2017}. Our main results are reported on the \texttt{test-dev} split for comparisons with state-of-the-art methods. Consistent with previous methods, the standard mask AP is used to evaluate the performance of instance segmentation. Moreover, we also report the box AP to show the object detection performance.

\subsection{Implementation Details}
ImageNet \cite{deng2009imagenet} pre-trained ResNet \cite{he2016deep} is employed as the backbone and multi-scale feature maps $\{F^l\}_{l=1}^{L}$ are extracted without FPN \cite{lin2017feature}.
Unless otherwise noted, the deformable attention has 8 attention heads, and the number of sampling points is set as 4. The feature channels in the encoder and decoder are 256, and the hidden dim of FFNs is 1024. We train our model with Adam optimizer \cite{kingma2015adam} with base learning rate of $2.0\times10^{-4}$, momentum of 0.9 and weight decay of $1.0\times10^{-4}$. Models are trained for 50 epochs, and the initial learning rate is decayed at $40^{th}$ epoch by a factor of 0.1. Multi-scale training is adopted, where the shorter side is randomly chosen within [480, 800] and the longer side is less or equal to 1333. When testing, the input image is resized to have the shorter side being 800 and the longer side less or equal 1333. All experiments are conducted on 16 NVIDIA Tesla V100 GPUs with a total batch size of 32.

\begin{figure}[t]
  \centering
  \includegraphics[width=1.0\columnwidth]{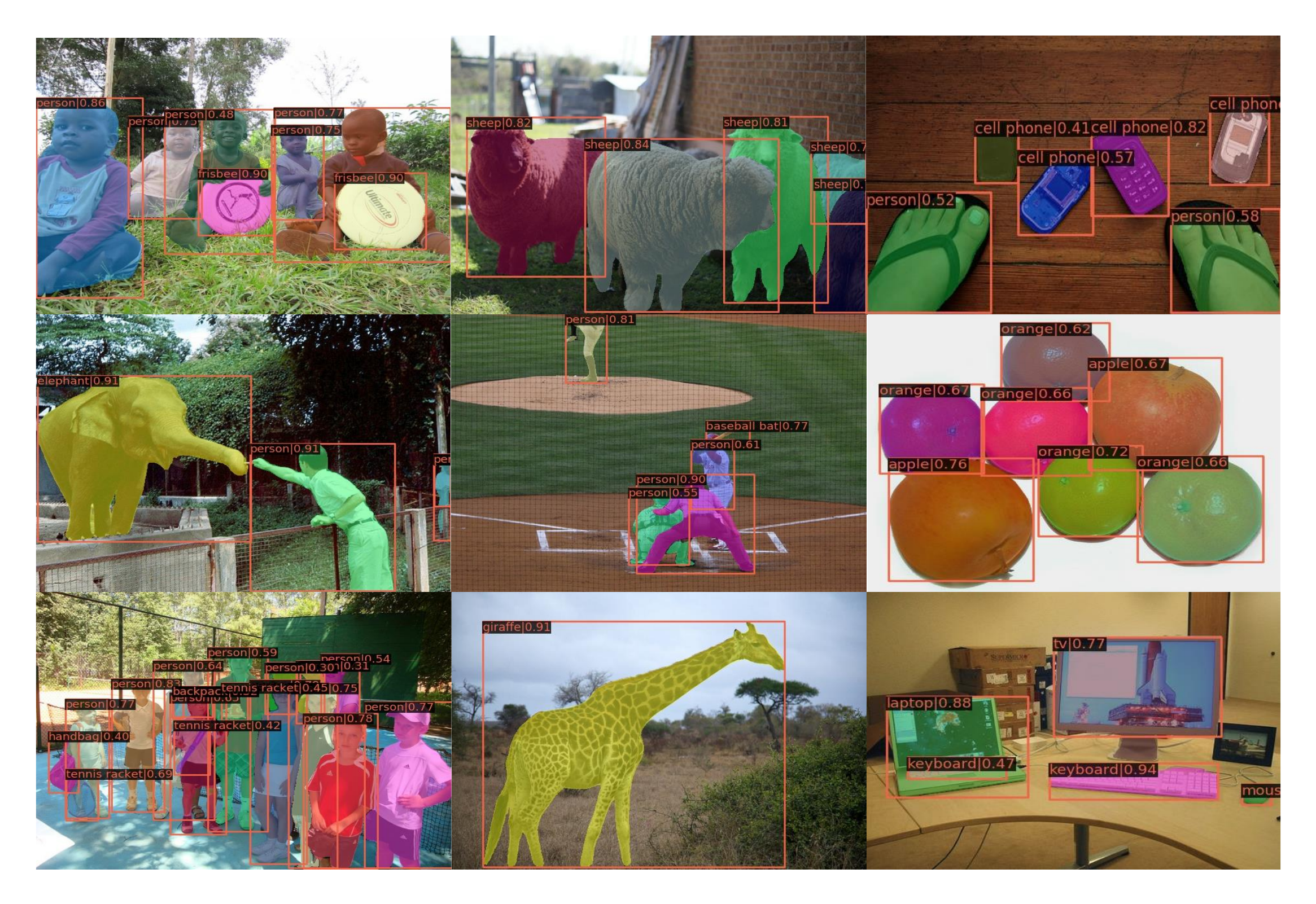}
  \caption{Qualitative results of object detection and instance segmentation on COCO \texttt{val2017} split. The model is trained on COCO \texttt{train2017} split with ResNet-50 backbone.}
  \label{fig:visualizatione}
\end{figure}

\subsection{Main Results}
As shown in Table \ref{tab:comparison with SOTA}, we compare SOIT with state-of-the-art instance segmentation methods on COCO \texttt{test-dev} split.
Without bells and whistles, our method achieves the best performance on object detection and instance segmentation.
Compared to the typical two-stage method Mask R-CNN \cite{he2017mask}, SOIT with ResNet-50 significantly improves box AP and mask AP by 7.8\% and 5.0\%, respectively.
The performance of SOIT is also better than the well-tuned HTC \cite{chen2019hybrid} by 4.2\% box AP and 2.8\% mask AP, which is an improved version of Mask R-CNN presenting interleaved execution and complicated mask information flow.
CondInst \cite{tian2020conditional} is the latest state-of-the-art one-stage instance segmentation approach based on dynamic convolutions. 
SOIT with the same ResNet-50 backbone outperforms CondInst with 4.7\% mask AP.
With a stronger backbone, ResNet-101, SOIT still outperforms the state-of-the-art methods over 2.0\% mask AP.
Benefiting from the RoI-free scheme, our method with ResNet-50 surpasses the recent SOLQ~\cite{dong2021solq} and QueryInst~\cite{Fang_2021_ICCV} by 2.8\% and 1.9\%, respectively.
We also apply SOIT to the recent Swin Transformer backbone \cite{liu2021swin} without further modification, building a pure transformer-based instance segmentation framework.
Our model with Swin-L can achieve 56.9\% and 49.2\% in box AP and mask AP, respectively.
As shown in Fig \ref{fig:visualizatione}, we provide some qualitative results of SOIT with ResNet-50 backbone on COCO \texttt{val2017} split.
Our masks are generally of high quality (\emph{e.g.}, preserving more details at object boundaries), and the detected boxes are precise.

\subsection{Ablation Study}
\subsubsection{Number of Heads in Instance-Aware Transformers.}
The multi-head attention mechanism is of great importance for the transformer.
In this section, we discuss the effect of this design on our instance-aware transformer encoder.
We vary the number of heads of multi-head attention, and the performance of instance segmentation is shown in Table \ref{tab:different heads}.
We find that using only one head of attention already has a moderate capacity and leads to a qualified performance with 37.8\% mask AP.
The performance of instance segmentation improves gradually with the increased number of attention heads in the instance-aware transformer.
Besides, when the number of attention heads increases up to 8, segmentation performance does not improve further.
We assume there are two reasons for this saturation in performance. One is that 4 different spaces of representation are sufficient for distinguishing various instances.
The other reason is that predicting too many parameters (873 parameters) makes optimizing the mask branch difficult.
Therefore, we set the number of attention heads in the instance-aware transformer to 4 by default in the following experiments.

\begin{table}[t]
    \centering
    \resizebox{0.95\columnwidth}{!}{
    \begin{tabular}{c|c|cc|ccc}
        \toprule
        Heads & AP  & AP$_{50}$ & AP$_{75}$ & AP$_{S}$ & AP$_{M}$ & AP$_{L}$ \\
        \hline \hline
        1  &  37.8 & 61.6    & 39.5    & 18.1   & 41.1   & 57.6  \\
        2  &  38.1 & 61.9    & 39.9    & 18.5   & 41.3   & 58.1  \\
        4  &  \textbf{38.4}  & \textbf{62.0} & \textbf{40.1} & \textbf{18.6}  & 41.7 & \textbf{58.4}  \\
        8  &  38.3 & 62.0    & 40.1   & 18.4   & \textbf{41.9}  & 58.4 \\
        \bottomrule
    \end{tabular}}
    \caption{Instance segmentation results on COCO \texttt{val2017} split with different number of heads of multi-head attention in instance-aware transformer. The input feature channel (\emph{i.e.}, $C_{mask}$) is fixed to 8 by default.}
    \label{tab:different heads}
\end{table}

\begin{table*}[ht]
    \centering
    \resizebox{1.0\linewidth}{!}{
    \begin{subtable}[]{0.5\linewidth}
        \centering
        \begin{tabular}{c|c|cc|ccc}
            \toprule
            Channels & AP & AP$_{50}$ & AP$_{75}$ & AP$_{S}$ & AP$_{M}$ & AP$_{L}$ \\
            \hline \hline
            4  &  37.6  & 61.8 & 39.2 & 18.2  & 40.8 & 57.5\\
            8  &  \textbf{38.4}  & \textbf{62.0} & \textbf{40.1} & \textbf{18.6}  & \textbf{41.7} & \textbf{58.4}  \\
            16 &  38.3  & 62.0 & 40.0 & 18.5 & \textbf{41.7} & 58.3  \\
            \bottomrule
        \end{tabular}
        \caption{Vary the output channels of the mask encoder.}
        \label{tab:different output channels of mask encoder}
    \end{subtable}
    \quad
    \begin{subtable}[]{0.5\linewidth}
        \centering
        \begin{tabular}{c|c|cc|ccc}
            \toprule
            Layers & AP & AP$_{50}$ & AP$_{75}$ & AP$_{S}$ & AP$_{M}$ & AP$_{L}$ \\
            \hline \hline
            0  &  37.9  & 61.4 & 39.4 & 18.0  & 40.9 & 57.6  \\
            1  &  \textbf{38.4}  & \textbf{62.0} & \textbf{40.1} & \textbf{18.6}  & \textbf{41.7} & 58.4  \\
            2  &  38.4  & 61.9 & 40.1 & 18.5  & 41.6 & \textbf{58.6}  \\
            \bottomrule
        \end{tabular}
        \caption{Vary the layers of stacked mask encoder.}
        \label{tab:different stacked layers of mask encoder}
    \end{subtable}}
    \caption{Instance segmentation results on COCO \texttt{val2017} split with different architectures of the mask encoder. ``Channels'': the number of channels of the mask encoder's output. ``Layers'': the number of stacked mask encoder.}
    \label{tab:different architectures of mask encoder}
\end{table*}

\begin{table}[t]
    \centering
    \resizebox{0.95\columnwidth}{!}{
    \begin{tabular}{c|c|cc|ccc}
        \toprule
        PE & AP & AP$_{50}$ & AP$_{75}$ & AP$_{S}$ & AP$_{M}$ & AP$_{L}$ \\
        \hline \hline
        None     & 37.9 & 61.4 & 39.6 & 18.3 & 41.2 & 58.0   \\
        Abs      & 38.4 & 62.0 & 40.1 & 18.6 & 41.7 & 58.4  \\
        Rel & \textbf{39.2} & \textbf{62.9} & \textbf{41.3} & \textbf{19.7} & \textbf{43.0} & \textbf{59.2}  \\
        \bottomrule
    \end{tabular}}
    \caption{Impact of the positional encoding in instance-aware transformer on COCO \texttt{val2017} split. ``None'' means  removing positional encoding, ``Abs'' represents the traditional absolute positional encoding and ``Rel'' represents the proposed relative positional encoding.}
    \label{tab:position encoding}
\end{table}

\subsubsection{Architectures of Mask Encoder.}
We then investigate the impact of the proposed mask encoder with different architectures.
We first change $C_{mask}$, \emph{i.e.}, the number of channels of the mask encoder's output feature maps (\emph{i.e.}, $F_{mask}$).
As shown in Table \ref{tab:different output channels of mask encoder}, the performance drops 0.8\% in mask AP (from 38.4\% to 37.6\%) when the channel of $F_{mask}$ shrinks from 8 to 4.
In this case, the multi-heads attention only has a single-channel map in each attention head.
It is hard for the attention module to obtain sufficient information on each instance.
Besides, the performance keeps almost the same when $C_{mask}$ increases from 8 to 16.
Thus, we fix the mask feature channels to 8 in all other experiments by default.
As the $C_{mask}=8$ and the number of attention heads is 4, there are a total of 441 parameters predicted by the mask branch for constructing the instance-aware transformer.

To demonstrate the effectiveness of the mask encoder, we directly connect a linear projection (output channel is 8) with layer normalization to the feature map $P_3$ instead of the proposed mask encoder.
As shown in Table \ref{tab:different stacked layers of mask encoder}, the segmentation performance drops 0.5\% (from 38.4\% to 37.9\%).
This result proves the importance of the mask encoder, which generates the specialized mask feature and decouples it from the generic image context feature.
Moreover, when more mask encoders are stacked, no noticeable improvement of performance is obtained, as shown in Table \ref{tab:different stacked layers of mask encoder} (3rd row).
This indicates that one mask encoder is sufficient, resulting in a compact instance segmentation model.

\subsubsection{Relative Positional Encodings.}
We further investigate the effect of our proposed relative positional encodings for the instance-aware transformers.
\emph{Abs} is the absolute positional encodings used in many transformer-based architectures \cite{carion2020end, zhu2021deformable}. \emph{Rel} is the proposed relative positional encodings in Equation \eqref{RelPosEncoding}, which employ the box center coordinates of object queries to obtain the instance-aware location information.
As shown in Table \ref{tab:position encoding} (1st row), the performance of our model drops 0.5\% in mask AP after removing absolute positional encodings to the mask features.
The instance-aware transformer cannot distinguish the instances with similar appearances at different locations without the positional information.
As shown in Table \ref{tab:position encoding} (3rd row), the relative positional encodings improve the segmentation performance of our SOIT by 0.8\% compared to the absolute positional encodings.
We argue that the relative positional encoding is highly correlated with the corresponding object query and provides a strong location cue, for instance mask prediction.
Therefore, in the sequel, we use the proposed relative positional encoding for all the following experiments.

\begin{table}[t]
    \centering
    \resizebox{1.0\columnwidth}{!}{
    \begin{tabular}{c|ccc|ccc}
        \toprule
        Stages & AP & AP$_{50}$ & AP$_{75}$ & AP$^{box}$ & AP$^{box}_{50}$ & AP$^{box}_{75}$ \\
        \hline \hline
        0  &  -   &  -    & -      & 46.8   & 66.3  &  50.7 \\
        1  & 39.2 & 62.9  & 41.3   & 47.3   & 66.2  &  52.0  \\
        2  & 40.7 & 63.6  & 43.4   & 47.6   & 66.4  &  52.5  \\
        3  & 41.2 & 63.9  & 44.1   & 48.1   & 66.5  &  52.8  \\
        4  & 41.7 & 64.2  & 44.5   & 48.2   & 66.4  &  53.0  \\
        5  & 42.0 & 64.5  & 44.9   & 48.5   & 66.7  &  53.2 \\
        6  & \textbf{42.2} & \textbf{64.6} & \textbf{45.3} & \textbf{48.9} & \textbf{67.0} & \textbf{53.4} \\
        \bottomrule
    \end{tabular}}
    \caption{Ablation of the number of decode stages enabling mask loss on COCO \texttt{val2017} split. Stages is $K$ means that enable the last $K$ decoder layers with mask loss. 0 stages represents a object detection model without any mask supervision. AP$^{box}$ denotes box AP.}
    \label{tab:different loss stage}
\end{table}

\subsubsection{Stages Enabling Mask Loss.}
Ultimately, we ablate the impact of the number of decoder stages enabling mask loss in training.
The classification and localization loss are enabled in all decoder stages in these ablations by default.
Note that we throw away all the predicted mask parameters in the intermediate stages when the training is completed and only use the final stage predictions for inference.
As shown in Table \ref{tab:different loss stage}, enabling more decoder layers with mask loss can improve both instance segmentation and object detection performance consistently.
The experimental results show that adding mask loss on all decoders can improve 3.0\% mask AP and 1.6\% box AP compared to enabling mask loss on only one decoder, respectively.
The gain of detection performance is mainly derived from the joint training with instance segmentation.
As shown in Table \ref{tab:different loss stage} (last row), the detection performance of the SOIT surpasses the pure object detector by 2.1\% (from 46.8\% to 48.9\%) with all decoder stages enabling mask loss.
This indicates the advantages of our framework, which learns a unified query embedding to perform instance segmentation and object detection simultaneously.

%% file: parts/5_conclusion.tex
\section{Conclusion}

In this paper, we present a transformer-based instance segmentation approach, termed SOIT.
It reformulates instance segmentation as a direct set prediction problem and builds a fully end-to-end framework.
SOIT is naturally RoI-free and NMS-free, avoiding many hand-crafted operations involved in previous instance segmentation methods.
Extensive experiments on the MS COCO dataset show that SOIT achieves state-of-the-art performance in instance segmentation as well as object detection.
We hope that our simple end-to-end framework could serve as a strong baseline for instance-level perception.

\section{Acknowledgments}
This work is funded by National Natural Science Foundation of China under Grant No. 62006183, National Key Research and Development Project of China under Grant No. 2020AAA0105600, China Postdoctoral Science Foundation under Grant No. 2020M683489, and the Fundamental Research Funds for the Central Universities under Grant No. xhj032021017-04 and xzy012020013.